# Reacting like Humans: Incorporating Intrinsic Human Behaviors into NAO through Sound-Based Reactions to Fearful and Shocking Events for Enhanced Sociability

Ali Ghadami[1, †], Mohammadreza Taghimohammadi[1, †], Mohammad Mohammadzadeh[1], Mohammad Hosseinipour[2], Alireza Taheri[1, *]

[1] *Department of Mechanical Engineering, Sharif University of Technology, Tehran, Iran*
[2] *Department of Computer Science, University of Padova, Padova, Italy*

**Abstract**

Robots' acceptability among humans and their sociability can be significantly enhanced by incorporating human-like reactions. Humans can react to environmental events very quickly and without thinking. An instance where humans show natural reactions is when they encounter a sudden and loud sound that startles or frightens them. During such moments, individuals may instinctively move their hands, turn toward the origin of the sound, and try to determine the event's cause. This inherent behavior motivated us to explore this less-studied part of social robotics. In this work, a multi-modal system composed of an action generator, sound classifier, and YOLO object detector was designed to sense the environment and, in the presence of sudden loud sounds, show natural human fear reactions; and finally, locate the fear-causing sound source in the environment. These unique and valid generated motions and inferences could imitate intrinsic human reactions and enhance the sociability of robots. For motion generation, a model based on LSTM and MDN networks was proposed to synthesize various motions. Also, in the case of sound detection, a transfer learning model was preferred that used the spectrogram of the sound signals as its input. After developing individual models for sound detection, motion generation, and image recognition, they were integrated into a comprehensive "fear" module implemented on the NAO robot. Finally, the fear module was tested in practical application and two groups of experts and non-experts (in the robotics area) filled out a questionnaire to evaluate the performance of the robot. We indicated that the proposed module could convince the participants that the Nao robot acts and reasons like a human when a sudden and loud sound is in the robot's peripheral environment, and additionally showed that non-experts have higher expectations about social robots and their performance. Given our promising results, this preliminary exploratory research provides a fresh perspective on social robotics and could be a starting point for modeling intrinsic human behaviors and emotions in robots.

## 1. Introduction

These days, thanks to the rapid growth of robotic technology, the usage of robots has broadened in healthcare [1], industry [2], consultation [3], and a wide variety of other tasks they can perform. One of the most exciting examples of that is the humanoid robots. The vital property of these robots is their ability to interact with humans in a natural manner. They can act as friends [4], teachers [5], playmates for kids [6], and in general, perform like a real human among the people. The physical appearance and acceptance of robots have recently received considerable attention in the field of human-robot interaction (HRI) [7]. While the appearance of robots significantly affects their acceptability, their behaviors also play a critical role in determining their level of acceptance among humans. However, this does not happen in the absence of human-like behaviors in the robot's conduct. Conventional human treatments and emotions arise from their instincts; for example, fear, upset, rage, happiness, and acts that are done unconsciously. Therefore, researchers have investigated numerous research to find out how they can make robots imitate human movements, hearing, and vision in different situations. Many equipment and methods have been developed, but one of the most used tools in this field is machine learning (ML). Machine learning models, especially deep models, can give robots the capability to show proper reactions according to the incidents occurring around them. In continuing, brief explorations are conducted into investigations on generating robot movements, as well as their vision and hearing abilities, to perceive their surroundings.

† *Authors contributed equally to the manuscript.*

*Corresponding author: artaheri@sharif.edu , Tel: +982166165531*

*1.1. Robot movement generation*

Movement in robots can be understood as functional and expressive movements [8]. The functional movements are designed to meet the designer's purpose in inventing the robot, and the expressive activities are a set of actions that mirror the delicate acts that humans usually do. Expressive movements can be accompanied by Ekman's basic emotions (i.e. fear, anger, joy, sadness, contempt, disgust, and surprise) [9], fatigue, mood, and other actions that are unnecessary for reaching the robot's goal. These actions can improve robot engagement with humans. For instance, investigations show that gazing during speech can enhance the interactions between robots and humans [10].

Numerous studies are being carried out in the field of generating movements for robots. Movement generation can be done by defining some degrees of freedom (DOF) for a robot model and then devoting a rational movement to them. Coulson simplified the human body to a model with six rotational joints and six DOFs [11]. In addition, to consider the body's general shape in this model, the center of mass was added as another variable. In the end, their proposed model had seven variables, for which they assigned some logical values to create postures attributed to anger, disgust, fear, happiness, sadness, and surprise. Nonetheless, this method can produce some misleading movements, so in this article, the results were evaluated with some observers as discriminators. Erden chose the best postures that Coulson [11] found, which are the postures that gain the best assessment, and implemented them on a NAO robot [12]. Obviously, this human movement generation method has some limitations. For example, it can lead to repetitive postures, which is undesirable because human movement naturally has always been accompanied by randomness. Therefore, the researchers have resorted to using deep ML models to address this problem.

Two main methods to generate movement for robots using ML models are recurrent neural network (RNN) and generative adversarial network (GAN) models. Zhao et al. used Long Short-Term Memory units (LSTM) to generate motion for a Baxter robot [13]. The input of the LSTM model was the palm and elbow positions of the human hands, which were collected by an RGB camera installed on the robot. The extracted features from the LSTM model entered into a fully connected layer to convert them into joint angles of the robot. In another work, Ochi et al. utilized an LSTM unit with an encoder and decoder model [14]. The proposed model got the sequential snapshots, which were taken by a Kinect camera installed on the robot alongside the robot's joint angles, and output synthesized motion to create the next step of the robot action.

In addition to using images as generator inputs, there are other methods to collect and generate data on favorable activity positions. Chen et al. used an armband equipped with EMG sensors and a hierarchical Bayesian model to create a collaborative system with a Baxter robot [15]. Mohammadzadeh et al. used a number of IMU sensors on wrists and thighs to collect data on some activities [16]. They designed a conditional GAN model to create synthesized data that resembled the actual obtained velocities and accelerations from the IMU signals. In these kinds of methods, the generated data can be converted to some positions on the human body, and due to the differences between human and humanoid robot physical dimensions, the obtained data must be mapped on the robot. Shahverdi and Tale Masouleh proposed a method that could map the human arm positions to NAO robot arms by solving the inverse kinematics of the robot and attributing the NAO robot size to human ones [17].

*1.2. Vision perception in robots*

Humans use their eyes to understand their environment; how and where different objects are located is essential for interacting with their surroundings. This demand necessitates a detector that detects and localizes the place of objects in the environment. To reach this desire, two main kinds of object detector models have been developed; one-stage and two-stage detectors. The two-stage detectors usually showed better precision in classification and localization, but with the advent of a one-stage detector named You Only Look Once (YOLO), the two-stage detector's usage declined due to the YOLO's efficiency compared to the two-stage detectors [18]. Zhang et al. utilized YOLOv3 to identify and locate their desired objects and subsequently grasp them using a NAO robot [19]. Li et al. implemented a modified YOLO model on a NAO robot to identify humans and provide guidance based on their mask-wearing condition [20]. Singh et al. developed a robot called Tinku to serve as a therapist and teacher for autistic children [21]. This robot uses Yolov3 and a single shot detector (SSD) for obstacle avoidance and expressing emotions.

*1.3. Auditory perception in robots*

Alongside vision, humans hear sounds and extract meaningful content to interact with their surroundings [22]. Robots need to react properly against different sounds; therefore, sound classification for interacting with surroundings is crucial for a social robot. As a result, humanoid robots must have this sense to understand the environment better. The footprint of sound detection can be found in the medical [23], voice assistant [24], and automotive industry [25] systems. Poore et al. implemented the frequency/band-pass filter and logistic regression methods to detect a predefined whistle [26]. They employed a NAO robot and showed that it could recognize the whistle sound using their proposed approach. Zheng et al. proposed a model based on a convolutional neural network (CNN) combined with random forest (RF) that is responsible for detecting emotions in human sound [27]. In this work, the sound signal was converted to a normalized spectrogram and given as input to the CNN model. The extracted features from the signals were used in the RF model for classification. Tripathi and Mishra tried to design an attention module along with a residual network to identify the most profitable segment of the spectrogram of sound signals and classify them sequentially [28].

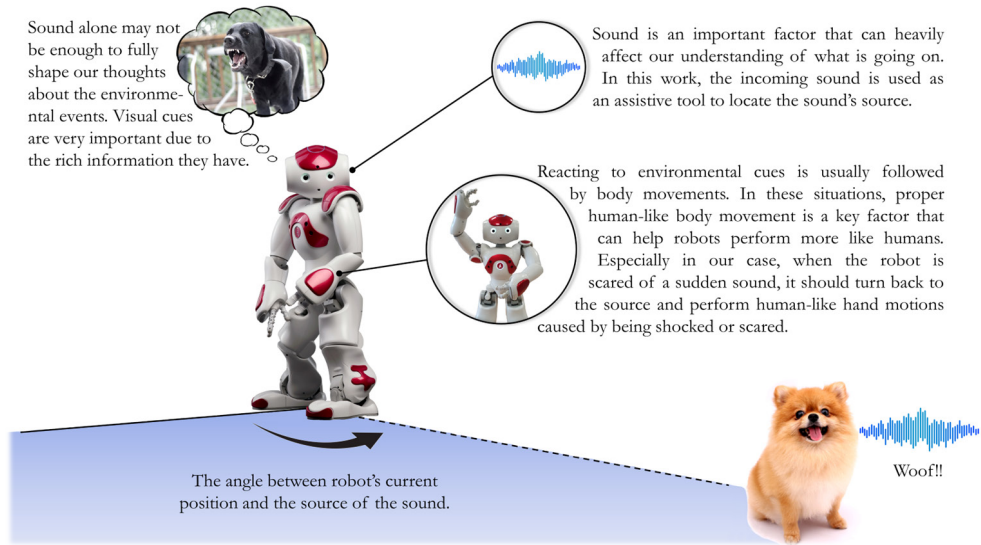

Figure 1. A scenario, where the robot encounters a sudden loud sound. A human-like reaction helps it to perform naturally while using both sound classification and object detection can perfectly locate the source of the sound.

*1.4. Integrated systems and scope of this study*

Although there have been many investigations in the past few decades on imitating individual human senses in robots, the number of articles that integrated them is limited. A number of works have focused on environment-aware expressive motion generation. In [29], speech-driven robot gestures by the generative adversarial network (GAN) approach were applied to the social robots, and its natural behavior effectiveness was evaluated. Also, in [30], a learning-based co-speech gesture generation model was explored to produce human-like gesture and speech content on the NAO robot.

It is evident that humanoid robots must possess the mentioned senses to communicate effectively and be more acceptable among humans. For instance, humanoid robots have become very popular as children's toys because they can effectively teach, play, and interact with kids. This has made the robots a preferred choice for parents, and sometimes even teachers and healthcare professionals to use them in teaching and healthcare applications. Therefore, it is necessary to make humanoid robots as much like a companion for children as possible by giving them innate human behaviors. The application of robots in the case of children is only one of the examples that we can count for humanoid robots today and it can justify the necessity of conducting more research to make robots more like humans, surpassing the traditional perceptions of robots in people's minds.

Previous studies have focused less on integrating senses and reactions that humanoid robots should have to simulate human unconscious behavior specifically their emotions and intrinsic reactions. However, to the best of our knowledge, these investigations have predominantly concentrated on individual senses that robots can emulate. It is important to note that relying on a single sense may not be convincing enough in replicating human-like behavior.

Creating robots exactly like humans is a long journey for researchers. However, it can be achieved by instilling human-like behavior in humanoid robots, one step at a time. This preliminary exploratory study aimed to introduce a human-like response in robots when confronted with fear induced by sudden and loud sounds as a step towards enhancing the robot's social cognition. By adding these human-like reactions, we hope to make robots more accepted by society. This study is just a small part of our larger effort to make robots act in ways that feel natural. We believe that the more robots act like humans, especially in scenarios where humans react intrinsically, the more people will be comfortable with them. This will not only make robots technically smart but also good at social interactions, making them more likable and relatable.

For this purpose, the generation of unique human-like motion was accomplished through the use of RNN-based networks. An audio classifier was employed to identify the type of sounds, while the YOLOv5 model determined the location of the fear-causing source (as depicted in Fig. 1). This multi-modal system was implemented on a NAO robot. Based on the survey results that was conducted in this study to assess the similarity of the robot's reactions to that of humans and the effect of the designed fear module on the robot's acceptability, it was concluded that this module makes the robot more intelligent and acceptable to humans.

The contributions of this paper can be summarized as follows:

(1) Video recordings from a predetermined number of scenarios in a laboratory setting, where individuals expressed surprise reactions upon hearing frightening sounds, such as loud and sudden noises were collected. All participants were aware of the scenario and intentionally performed the reactions.
(2) A multimodal system that imitates the reaction of humans to sudden and loud sounds composed of three ML-based models for human-like unique motion generation, sound classification, and object detection was trained and implemented on NAO.
(3) Two groups of experts and nonexperts in robotics participants were surveyed to evaluate the robot's performance when encountering loud and sudden sounds typically accompanied by human reactions in that environment.

The subsequent sections of the paper are structured as follows: Section 2 outlines the methodology employed in this study, encompassing the deep learning tools utilized, the datasets accessed, real-world implementation details, and the assessment tools employed. Section 3 delves into the results of the networks' performance, the systems' outputs, and implementation, alongside the findings from user studies. Sections 4 and 5 offer discussions on the results and limitations, respectively. Lastly, Section 6 presents the conclusion drawn from the study.

## 2. Methods

In this section, the methodology and implemented tools are described in detail. This study contains three modules: sound classification, motion generation, and object detection models. The initial step towards the purpose of the paper, meaning reacting properly to a sound, is to recognize the sound's type (sound recognition module) and its direction, where the latter can be determined using an array of microphones of the NAO robot. When the robot detects a loud and sudden sound that could trigger fear or a threat response, it will act a human-like action (i.e., raising hands), which in this study is limited to arms movements (action generation module). Finally, when the robot reacts and turns to the direction of the sound, it will detect the fear-causing source using a combined sound detection system and object detection system (YOLO), which will be explained later. The general pipeline of the work as described can be found in Fig. 2. In the following, the details of the implementations of each module are described.

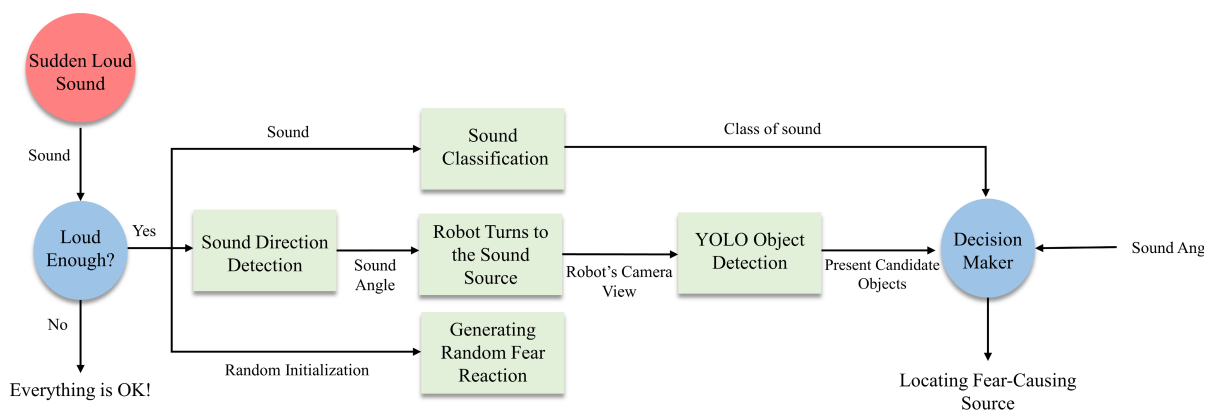

Figure 2. The complete pipeline of the fear module.

*2.1. Datasets*

The dataset used for human motion generation was collected by an RGB camera for seven different types of fear actions from two subjects, and each class contains 40 videos between 4 to 7 seconds. Subjects were asked to perform fear-inspired actions multiple times to make the model capable of generating diverse actions. Some samples of this dataset are shown in Fig. 3, while the complete set of our dataset classes can be found in Fig. 3.

The audio dataset consists of 1 to 5-second-long recordings organized into 14 classes with 40 examples per class, including thunderstorm, glass breaking, siren, car horn, barking dog, door wood knock, door wood creaks, crying baby, sneezing, clapping, coughing, laughing, female scream, and male scream, collected from ESC-50 [31] and NIGENS [32] dataset. The reason for selecting these classes is that they consist of sounds that have the potential to evoke feelings of shock or fear in certain situations.

The dataset used to train YOLO is gathered through search engines like Google and manually adding some pictures of objects in the real-world environment. We tried to identify candidates of fear-causing sources in an indoor environment, and the final classes were chosen as dog, window, human, door, and television. Each class contains roughly 400 training samples. It is notable that we can relate every sound class to one or more of these classes as Table 1. Notably, this work is a proof of concept and we are considering only a small group of objects and scenarios to show the validity of the method.

*2.2. Motion Generation*

Human motions have many elaborations that are essential for humanoid robots to do them to be like real humans. It is impossible for robots to emulate human motions without an appreciation of the properties of human ones. Complete comprehension of human motion kinematics is complex, so it can be hard to generate human motion theoretically. One of the appropriate approaches for motion generation is applying sequential machine learning models. This study generated the arm movements in 3-dimensional coordinates using Long Short-Term Memory (LSTM) and Mixture Density Networks (MDN). LSTM model was preferred because of its ability to learn the relations of the motions in different moments and use its knowledge to create an instance motion like a real one. However, the LSTM model can limit the generated motions and is not a creative model; that's why combining this model with an MDN model is imperative. Using MDN models gives this chance to have more innovation in the synthesized data.

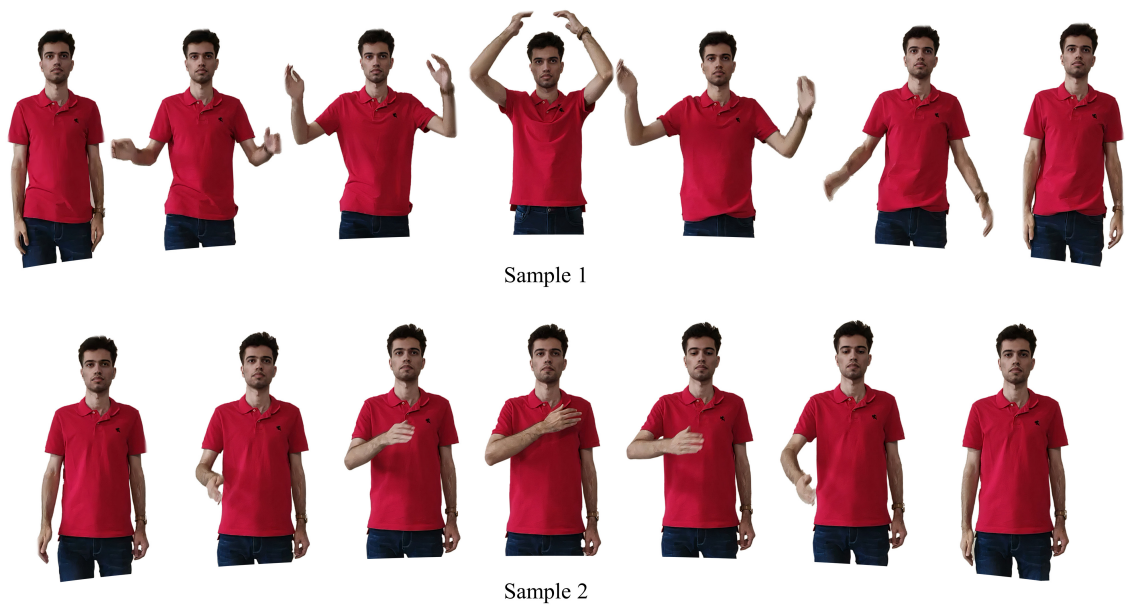

Sample 1

Sample 2

Figure 3. Two samples of our motion dataset.

Table 1. Relationship between sounds and objects of the dataset

| Visually detected object | Possible related sounds |
|---|---|
| Dog | barking, glass breaking |
| Window | thunderstorm, siren, car horn, female scream, and male scream, barking |
| Door | door wood knock, door wood creaks |
| Human | sneezing, clapping, coughing, laughing, female scream, male scream, glass breaking |
| TV | All of the sounds |

For creating the motions, choosing some specified fixed points on the arms is necessary. For this measure, the minimum number of points was selected to decline the need for more data for training the generator model. As a result, the selected points are placed on both arms' wrists, elbows, and shoulders (shoulder points are assumed to be fixed and were not used in the training process). The dataset collected by recording videos of seven actions and considered points' positions were extracted using MediaPipe from videos. MediaPipe Pose tracks body pose with high fidelity using 33 3D landmarks and a background segmentation mask from RGB video frames without manual intervention [33].

In addition to the inputs previously mentioned, a linear function was employed to determine the end of the motion sequence. This function starts at zero at the beginning of the motion and gradually increases to one at the end. The resulting value serves as the network's input and output (with one step shift to the right). In this paper, we referred to it as the progressive LSTM, a technique similar to progressive transformers which were previously used in [34]. This approach allows the network to monitor the progress of the motion sequence constantly.

MDN networks are the combination of a common neural network and a mixture model on top of that. As Fig. 4, MDN has three outputs for a single output of our network. The first output is the mixing coefficient ($\alpha$), which is the impact factor of each Gaussian distribution on the output. The second output is the mean of the predicted output ($\mu$), and the last output is the standard deviation of the predicted output ($\sigma$). As it is depicted, unlike conventional networks, these networks can predict means and standard deviation for every Gaussian function and thus can cover a range of possible solutions. This can help us generate unique actions each time we initialize the algorithm. The LSTM used in this work is a single layer of a bidirectional many-to-one network with 64 hidden units. The network's input is a vector of spatial coordinates of the arm joints, gathered by the MediaPipe in the last 20 frames and the linear function discussed before. The output will be the spatial coordinates and the value of the linear function of the next frame. In the inference time, the network's output will be fed to the input, and the new output will be generated with the new 20 frames (composed of the 19 frames from the initial condition and a newly generated output) and so on. Notably, a recurrent drop-out of 0.4 was used, and 15 Gaussian functions were chosen. The results are shown in the next section in detail.

## 2.3. YOLO Object Detection

In this work, the object detection task was done using YOLOv5. To recognize the objects, YOLO divides the frame into a grid of cells. Each cell is responsible for predicting K bounding boxes and their confidence scores instead of looking for the region of interest within the frame. Each bounding box contains four numbers (i.e. X-center, Y-center of the object, and its width and length) to help us locate the object within the frame. After deploying the algorithm on the frame, Threshold Filtering is used to filter the detected areas with low confidence. Also, Non-max Suppression will be performed to omit the extra bounding boxes of the same object and keep the bounding box with the highest confidence score. The notable feature of YOLO is detecting different objects with the same center points in one cell grid. For YOLO to pick multiple objects with the same center cell grids, it defines anchor boxes with different width/length ratios in order to detect multiple objects. YOLO was preferred in this work due to its high speed and accuracy.

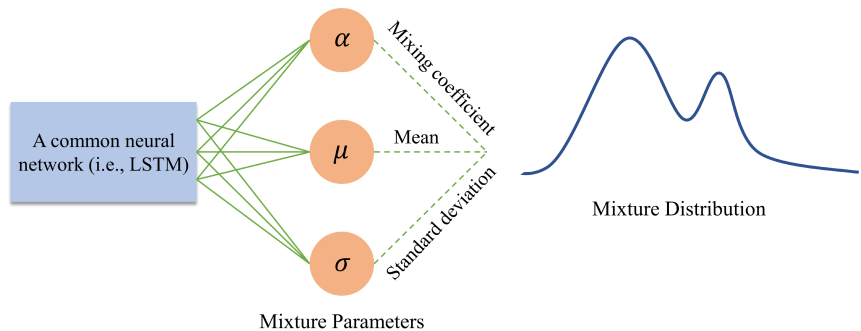

Figure 4. The MDN architecture.

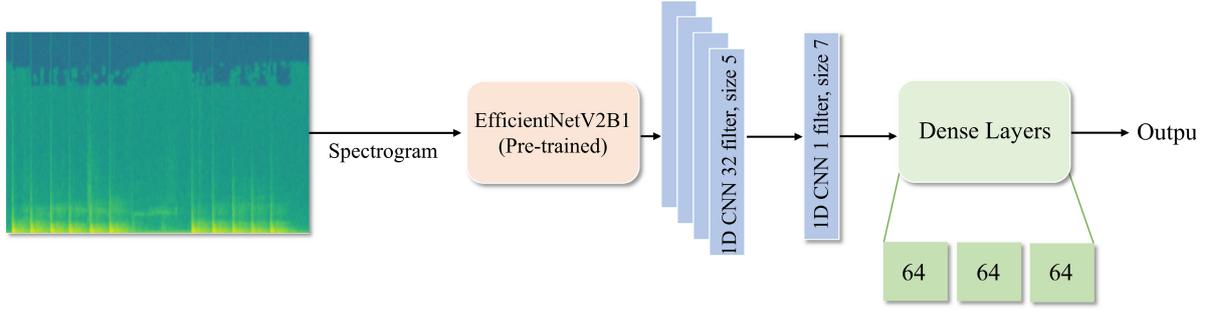

Figure 5. The sound classifier used in this study. The input was the spectrogram of the sound, and the EfficientnetV2B1 was used as a feature extractor.

Different versions of YOLOv5 are available. Among them, the YOLOv5-large was chosen and trained on the dataset. Also, 10 percent of the data were separated as the test data. The model was trained for 300 epochs and with an image size of 800×600.

*2.4. Sound Classifier*

The spectrogram of the input audio signals was calculated from the time-domain signal using the Fourier transform, as this data contains rich features both in the time and frequency domains. Transfer learning was preferred in this work for training the sound classifier network. EfficienNetV2B1 [35] was chosen as the feature extractor as it showed superior performance in classification using spectrograms [36]. In the following, 1D convolutional layers were used, and finally, dense layers were utilized to predict the correct class of the total 14 classes. The detailed architecture of the sound detection classifier is provided in Fig. 5. Convolutional activation functions were ReLU, and the dense layers had the LeakyReLU activation function. Also, Adam optimizer with a learning rate of 0.001 was used. The pre-trained network weights were not trainable, and only the rest of the network was trained.

*2.5. NAO Implementation*

This section contains two main parts. First, the mapping method used to map joint coordinates of human to the NAO motor angles will be introduced; and then, the implementation of the proposed multi-modal system (i.e., sound classification and object detection) will be described in detail.

*2.5.1. Joint Mapping*

As described before, shoulder, elbow, and wrist 3D positions were extracted from the data with MediaPipe. A mapping approach described in [17] was used to map these positions to the robot joints. The mapping equation, which is presented in Eq. (1), contains vectors from one joint to the next on both human and robot links, denoted by $\vec{V}_{Human}$ and $\vec{V}_{robot}$, respectively.

$$\frac{\vec{V}_{Human_i}}{|\vec{V}_{Human_i}|} = \frac{\vec{V}_{robot_i}}{|\vec{V}_{robot_i}|} \rightarrow \vec{V}_{robot_i} = \frac{|\vec{V}_{robot_i}|}{|\vec{V}_{Human_i}|} \vec{V}_{Human_i} \qquad (1)$$

In Eq. (1), the robot vector can be utilized as the kinematics operation of its arms. For obtaining angles of the robot motors for action generation, the obtained vectors were used in an inverse kinematics procedure. For this aim, first Denavit Hartenberg (DH) convention was implemented by assigning the link frames for each arm. Then, forward kinematics was calculated by representing the angles of the robot's arm for obtaining the position of one joint to the prior one. Finally, the angles of the robot joints were derived by equalizing the vector derived in Eq. (1) and forward kinematics solution. Notably, the mechanical limitation of the NAO robot has been considered in this algorithm for collision avoidance.

*2.5.2. Experimental Setup*

The NAO robot was placed in a pre-designed social environment. As soon as a sudden and loud sound occurred, the robot had to determine where the sound came from and which object in the environment could be the source. Additionally, the robot's reaction to the sound by moving hands and turning to the source of the sound was of interest to us.

For performing the selected scenario, the robot utilized three commands in NAO python library: "Sound Detection", "Sound Localization", and "Audio Recording". "Sound Detection" was used to identify the occurrence of a loud sound. By using the "Sound Localization" command, the robot could determine the appropriate angle for turning towards the sound. Also, the robot recorded sound simultaneously using the "Audio Recording" command. During the scenario, audio was recorded using the robot's microphone, and the time of sound detection was used to specify the exact moment of the event occurrence. Furthermore, environmental noise (containing the robot's fan noise) was removed before sending the audio signal to the sound classification model.

In response to the sound, the robot should rotate toward it, initiating the motion generation phase. As part of this phase, the NAO robot is expected to exhibit human-like behaviors when experiencing fear or shock, making the robot's reaction more natural and convincing. The robot's camera in its head takes three pictures after it completes the rotation and reaches the desired angle. First, a photo will be taken from the detected sound angle, whereas the other two will be taken at -15 and +15 degrees from the robot's straight position. It is later determined which object made the sound by using these pictures.

In the following, the photos and recorded audio are sent to two different models: the sound classifier and the YOLO object detector. These two models were developed to identify which objects were probably causing the sounds based on the pictures and audio. Comparing outputs from these two models, the robot could determine which object in the environment is most likely to be the sound source.

Overall, the robot in this scenario can be programmed to respond to specific stimuli in its environment, as well as to use advanced technologies, such as sound detection and object detection, to fulfill complex tasks. More specifically, NAO will decide on the source of the sound based on three metrics: probabilities of the detected sounds, the confidence of the YOLO object detector for detected objects, and the normalized distance of the object detected by YOLO from the sound direction. These parameters will be multiplied together for each possible sequence of parameters as Eq. 2 and the event with the most score will be chosen as the source of the sound.

$$Score_i = (sound\ probability)_i \times (YOLO\ confidence)_i \times (1 - distance)_i, \qquad (2)$$

Where "i" in this equation corresponds to the valid (sound, object) pairs (refer to Table 1) from detected sounds and objects.

*2.6. Assessment tools*

For evaluation of the research, in addition to the accuracies of each sub-part of our multi-modal system, the opinion of the people was asked about the results and performance of the robot. Three videos were prepared for three different scenarios, where the robot encounters a sudden and loud sound and has to react properly using our proposed system. The scenarios were selected to trigger various aspects of our algorithm and emulate real-world events as much as possible. In each video, Nao will react to sound as discussed before, by turning to the source of the sound while moving its arms, and finally locating the source of the sound by saying the name of the object. Each video lasts approximately 40 seconds. A survey was designed to ask the participants to answer ten questions by giving a score between one and five. The questions were chosen so that all aspects of this research can be evaluated. Questions 1 to 7 were designed by the authors because these questions, intended to investigate the hypotheses of the research and robot's performance, were not found in the UTAUT questionnaire, SP (Social Presence) items [37]. Additionally, questions 8, 9, and 10 were chosen from the UTAUT questionnaire, SP items [37]. These questions contain:

- Q1: How much do you consider the behavior of the robot to be similar to the behavior of humans when afraid in similar situations?
- Q2: How much has the robot's behavior caused you to feel insecure and afraid at the moment of fear?
- Q3: How successful do you think the robot's strategy was in finding the source of the sound?
- Q4: How close do you think the robot's strategy to find the source of the sound was to human behavior in similar situations?

- Q5: In your opinion, how much does this behavior of the robot increase the acceptance of this social robot in society?
- Q6: How much do you consider the robot's reaction to be fantasy or close to reality?
- Q7: How much will the robot's appropriate response to environmental stimuli (such as sounds, relationships between people, the atmosphere of the environment, etc.) increase your trust in the robot and make it accepted as a member of society?
- Q8: I can imagine that Nao is a living being.
- Q9: I often thought that Nao was not a real person.
- Q10: Sometimes it seemed that Nao had real feelings.

The participants should rate the items on a five-point Likert scale (from 1 to 5). The scale included verbal anchors ranging from "very low/totally disagree: 1" to "very much/totally agree: 5", allowing the subjects to express their opinions on the questions/items.

For the survey, two groups of expert and non-expert people in the social robotics field were chosen. Expert subjects include 35 Ph.D. and Master students researching robotics and AI-related topics, and non-expert subjects contain 35 people with no academic knowledge about robotics and AI-related topics. The participants watched videos and results via the online prepared survey and at the end, they were asked to submit their answers to the questions. Our decision to conduct an online survey rather than an in-person one (for example, showing everyone the robot's performance live and in person) was based on the fact that all participants could make their decision based on exactly the same content and conditions, as well as the fact that many of the participants could not make an in-person visit due to their location. Participants could go back and forth in the survey, and also rewatch a video several times. The overall process lasted approximately 10 minutes per participant.

We formulated the following hypotheses based on our expectations and previous related studies [38, 39] (H1 is related to the robot's performance and H2 is related to the expectations of people about a social robot):

H1. The proper reaction of the robot to sudden environmental sounds can enhance its acceptance among people, and it can induce in the viewer's mind that the robot is something more than codes and algorithms.

H2. Individuals lacking expertise in the field of robotics and AI tend to hold higher expectations regarding the performance of a social robot compared to those who possess expert knowledge in these fields.

Hence, the outcomes from our questionnaire aid in evaluating the algorithm's performance while also validating or challenging the stated hypotheses.

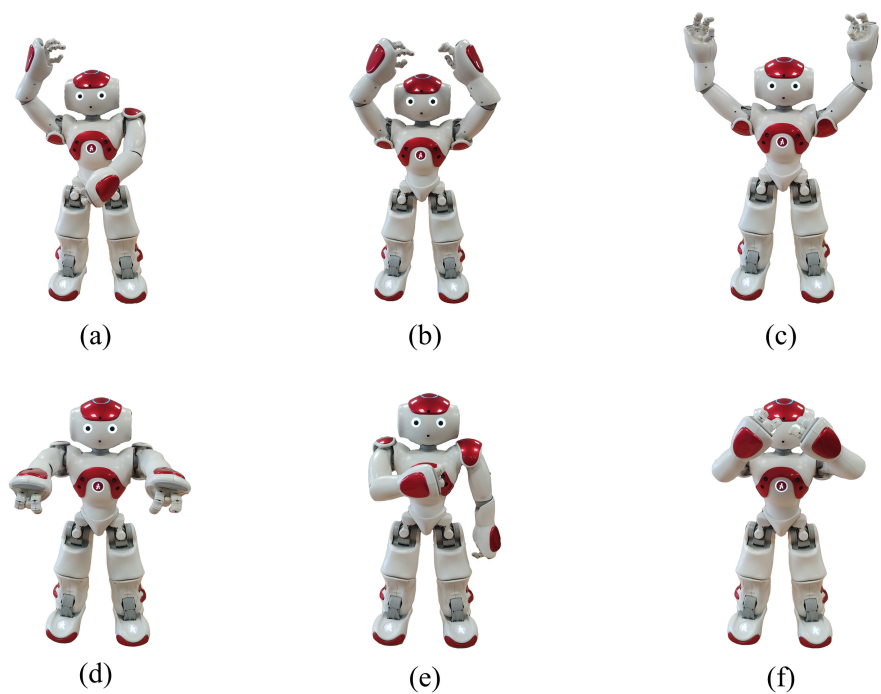

Figure 6. Selected hand motions, inspired by human intrinsic movements when shocked or scared by a sudden and loud sound.

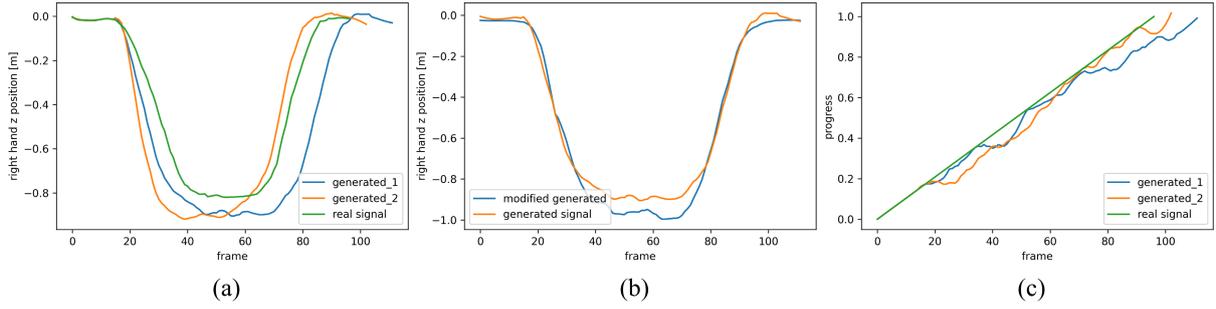

Figure 7. Results of the motion generation module for class (a) illustrated in Fig. 6. (a) generated and real captured data for the *z* position of the robot's right hand, (b) the effect of denoising and length modification on the positions, and (c) the linear function predicted by our network alongside with the ground truth data.

## 3. Results

This section provides and discusses the results of the described algorithms in section 2. Note that the datasets and codes are available upon logistic request.

*3.1. Motion Generation*

First, the results of the motion generation network are provided. Seven separate networks were trained for each of the seven action classes. A preview of the collected actions is shown in Fig. 6. These actions are chosen based on the natural human reactions when facing a similar situation.

After the networks were trained on our datasets, a length modification was done by forcing the length of the arm and forearm to be constant. Finally, to have a smooth trajectory, the data is denoised, i.e., if the distance of a specific point did not exceed a predefined threshold for some frames, the average coordinates of those frames were replaced for those frames. This will help dampen the noises. A sample of the results is provided in Fig. 7 and Fig. 8 for one of the motion classes. Notably, every time the network is run, we get a unique motion, just like humans! Choosing the type of fear and initializing the motion sequence can be selected randomly. Results of this part show a good performance from the networks, as they generate human-like and unique motion sequences.

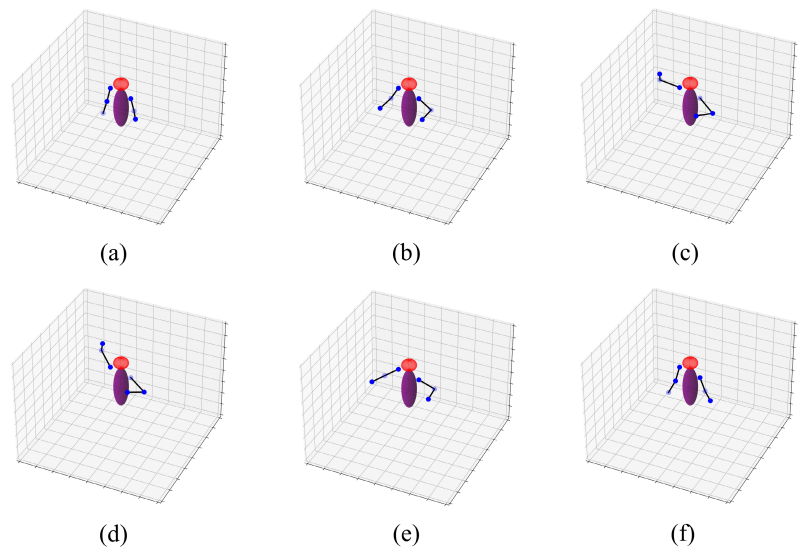

Figure 8. A sample of the generated motion from (a) to (f).

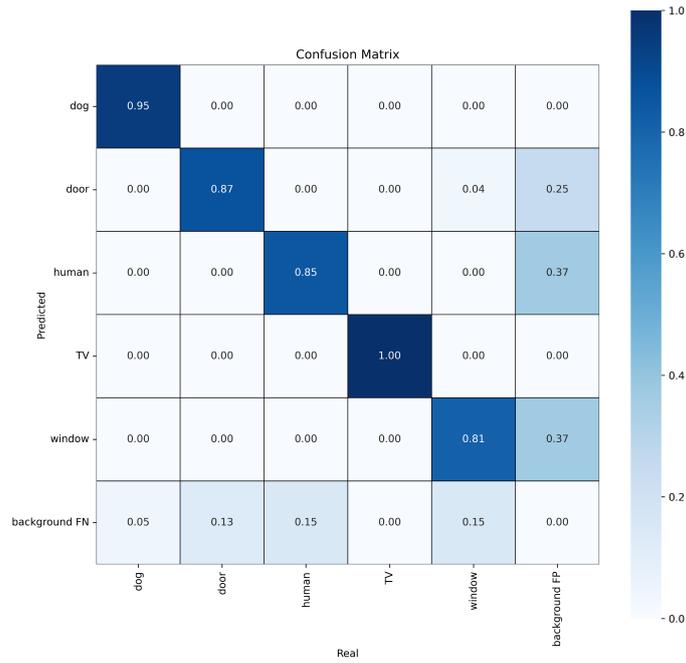

Figure 9. The confusion matrix for YOLO object detection.

*3.2. YOLO Object Detection and Sound Classification Models*

This part discusses the effectiveness and results of the trained YOLOv5 and sound classification models in detail. The YOLO model was trained for 300 epochs with MAP (Mean Average Precision) of 95% with 50% confidence, and 71% with 95% confidence for the test data. The confusion matrix of the test results is shown in Fig. 9. As depicted, we have a solid and acceptable performance for real-environment usage. According to the model's performance, it can be a promising candidate for object detection for NAO to find the probable candidates of the sound source.

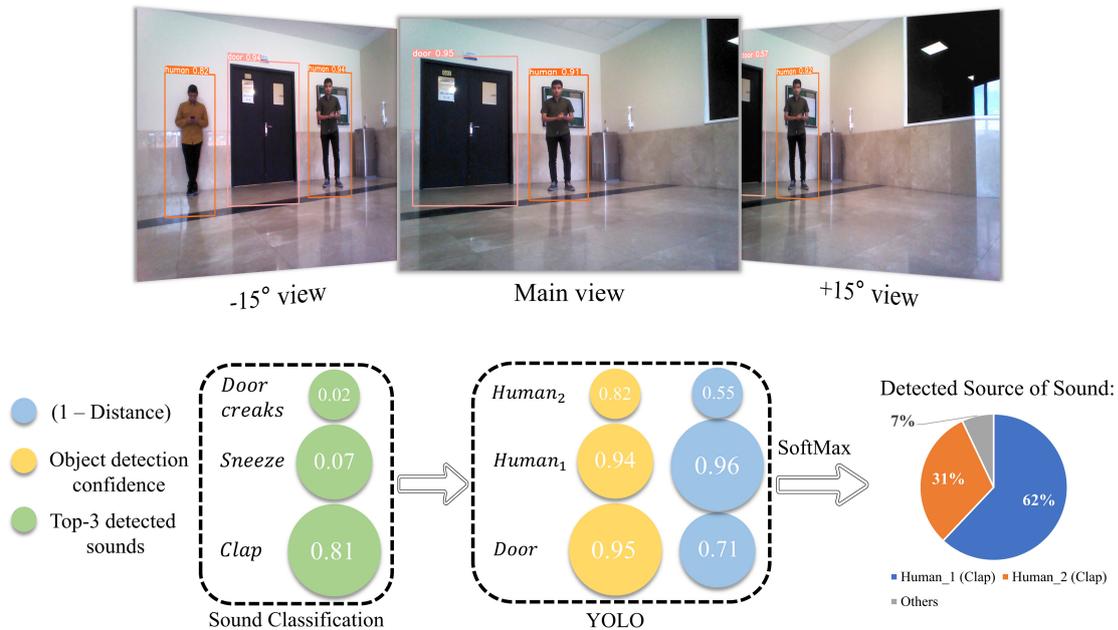

Figure 10. An example of conducted scenarios. In this example, the robot took three pictures when turned to the sound source and performed hand motions. The normalized distance of each detected object in images was calculated, and (1 – Distance) was used as a parameter to determine the source of the sound along with other parameters. The multiplication of the probabilities of the sound classes determined by the sound classifier, YOLO object detection confidence, and (1 – Distance) was used to determine the fear-causing source. (Others in the pie chart correspond to $human_1$ and $human_2$ Sneezing and door creaks).

Table 2. Average execution times and delays

| Event | Average execution time (s) |
|---|---|
| Robot's movement delay after sound occurrence | 0.86 |
| Taking pictures and saving (per picture) | 1.92 |
| YOLO object detection (per image) | 0.934 |
| Sound classification | 0.124 |
| Communication delay | 1.21 |

Sound detection is an essential task that can help localize the fear-causing source when multiple candidate objects are present in the robot's view. The final categorical accuracy of the model on the test data was 89.3%. It is evident that the network had a reasonable performance and can perfectly discriminate classes. The ability of the model to classify the incoming sounds along with the object detector model can help us toward localizing the source of the sound in the environment.

*3.3. Implementation Results*

The complete system was implemented on the NAO robot as the last step. Now the robot is expected to react naturally when facing a sudden and loud sound. More specifically, the robot should first detect the occurrence of a sound that can possibly cause fear or shock based on the intensity of the incoming sound. Then, while performing arm movements turns to the source of the sound. Lastly, utilizing the sound classification model and YOLO object detector, it should detect the source of the fear-causing sound in the environment. The average execution time of each part of the proposed system is provided in Table 2. As these delays are not occurring simultaneously, they do not have a huge impact on performance, but they are still a concern.

Humanoid robots and the research that is conducted on them are entirely done to make them plausible for human societies. For this aim, human opinions about the treatment of robots are essential. Therefore, the most dependable way to evaluate robot behavior is through a human assessment. In this way, in addition to the evaluation that is done by measuring the accuracy of desired robot applications, we attempted to collect the opinions of people regarding their perception of the robot's performance. Some scenarios were designed (i.e., Fig. 10), and the robot's reaction video was recorded. A total of three videos were selected to be shared with our subjects to evaluate the robot's behavior and to understand how successfully we prospered in making the robot more sociable, acceptable, and intelligent. The results of this survey are provided in Fig. 11 via boxplots.

Generally, the results of the survey indicate that we were successful in reaching the goal of this research, which was increasing the robot's sociability and acceptance by developing an algorithm for reacting to sudden sounds properly. Both expert and non-expert participants agree that the robot's behavior to some extent was close to that of humans, and the strategy was successful. The only point is non-expert participants, despite experts, slightly disagree with the statement that NAO was a living being and had real feelings. Still, they agree that this behavior (which in their mind does not make the robot real) has a positive effect on the acceptability of that in society. Thus, these results generally support the H1 hypothesis, which is in line with our expectations.

Table 3. The results of the T-test on the participants' survey. The scores are out of 5 and P-values less than 0.05 are shown in bold

| Question No. | Mean *(SD)* | | T-Value | P-Value |
|---|---|---|---|---|
| | **EXPERTS** | **NON-EXPERTS** | | |
| Q1 | 3.357 *(0.894)* | 3.167 *(0.957)* | 0.67 | 0.51 |
| Q2 | 2.070 *(0.960)* | 2.222 *(1.227)* | -0.45 | 0.65 |
| Q3 | 4.571 *(0.562)* | 4.056 *(0.848)* | 2.42 | **0.02** |
| Q4 | 3.964 *(0.865)* | 3.333 *(1.000)* | 2.21 | **0.03** |
| Q5 | 3.893 *(0.900)* | 3.556 *(1.065)* | 1.13 | 0.27 |
| Q6 | 2.75 *(1.022)* | 3.222 *(0.975)* | -1.52 | 0.14 |
| Q7 | 3.75 *(1.090)* | 3.278 *(0.990)* | 1.45 | 0.15 |
| Q8 | 3.107 *(1.205)* | 2.278 *(1.190)* | 2.236 | **0.03** |
| Q9 | 3.107 *(1.175)* | 3.778 *(1.181)* | -1.84 | 0.07 |
| Q10 | 3.250 *(1.271)* | 2.667 *(1.374)* | 1.44 | 0.16 |

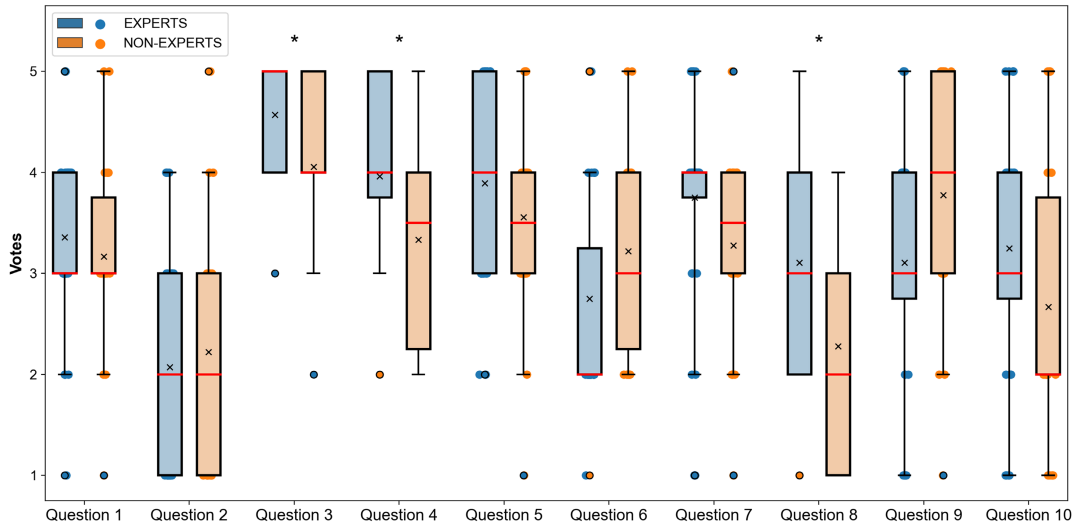

Figure 11. results of the implemented survey.

By performing T-tests on the results for subjects in the two groups (experts and non-experts), which is shown in Table 3, it is indicated that for questions 3, 4, and 8, the results between the two groups showed significant differences (p-value < 0.05). In questions 3 and 4, while both groups' scores are above average, expert people are more satisfied with the strategy of the robot for finding the source of sound. In question 8 however, which can reflect all of the aspects of the work, the score given by non-expert people is below 3, indicating that they are not convinced that the robot performed like a living being. In contrast, expert people believe that it somehow acted like a living being. Aside from these questions where the difference between expert and non-expert individuals was significant, in the remaining questions the expert participants had a more positive opinion about the results too. These findings can support our H2 hypothesis that can originate from the fact that the imaginations of non-expert people about social robots, or robots in general, are shaped by movies and the hype about AI, while expert people in this field treat that more realistically.

## 4. Discussion and Future Works

In this study, we investigated designing the fear instinct multi-modal module for a humanoid robot on NAO. This research contains three major parts; sound classification, object detection, and motion generation. This study aims to increase the sociability and acceptance of humanoid robots in societies by adding humans' natural behavior in response to environmental events to a humanoid robot. The fear module was practically implemented on the NAO robot, and the robot's capabilities were assessed individually and in combination with the presence of a frightening or shocking element. The conducted research and its results demonstrate several notable achievements.

There is a lot of independent research on each section of our study. In these works, the researchers tried to enhance only one sense of the robot: vision, motion, or hearing. But the importance of using these senses simultaneously is mainly neglected. In this research, we use two vital human senses, which means hearing and vision, in conjunction with proper movements to depict an intrinsic behavior of humans, i.e. fear, on the NAO robot. In this study, the designed multi-modal module was practically implemented on the NAO robot, and its behavior was evaluated.

Furthermore, the robot's reactions when it heard a sudden and loud sound varied from its previous reactions. This success was achieved using a motion generator to synthesize the action and preparing several pre-designed scared responses that were tailored for the robot as training data. This randomness in its behavior helped the robot to act as a real person in the same situation.

As a preliminary exploratory study, we examined how the robot can interact well with its surroundings and react to sudden changes in its environment focusing on one of Ekman's basic emotions (i.e. fear). To the best of our knowledge, due to the lack of similar works in this area, we cannot systematically compare this study with others; however, we investigate some studies for developing emotional postures [12, 13], expressing emotions [40] and detecting the target [20] for social robots. These works demonstrated the importance of imitating human senses for acceptability in social environments and multi-modalities (i.e. voice, body, and vision) to improve this concept.

Finally, as a survey inquiry, we collected people's opinions about the robot's behavior when confronted with a sudden and loud sound. The results showed that most participants agreed that the robot had a natural behavior in the presence of sound, which evoked this image in one's mind; the robot is aware of its environment, so it is more trustable. The results of the T-test showed that experts are aware of the challenges and limitations of implementing experimental setups on social robots and have fewer expectations than non-experts. It is in line with previous studies indicating that non-experts have high expectations for robots because of their experiences with fictional robots from media [39]. It is also similar to the results of [38], where university students were found to recognize the emotions generated by the Nao robot better than the other groups. Additionally, Tulli et al. suggest that there is a gap between the expectations of users and the capabilities of social robots, as users tend to have assumptions and expectations about robots that may not align with their actual abilities. On the other hand, researchers' expectations are influenced by their academic backgrounds and university priorities, which may differ from the expectations of users [41].Alongside the advantages mentioned in this study, some challenges remain that can be considered for future works. Although the proposed method can induct the fear instinct into the soulless robot's body, humans exhibit a wide range of habitual reactions, understandings, and behaviors without thinking about them. Therefore, it is conceivable that a significant amount of work remains before a robot can truly resemble a human. However, particularly regarding existing models in our proposed multimodal system, further research can be conducted.

In the case of movement generation, considering a broader range of fearful reactions could enhance the authenticity of the robot's responses. Additionally, various approaches can be explored to generate personalized reactions suited to different sounds. For instance, the sound of breaking glass might prompt the robot to instinctively protect its head with its hands. In such cases, zero-shot algorithms can be utilized to determine the most appropriate reaction for a specific sound.

The process of detecting the source of a sound sometimes involves experiencing unfamiliar objects that produce sound, which individuals become familiar with over time. Therefore, a potential future direction for research is to make robots more realistic by giving them a continual learning model in their auditory and visual systems, in addition to what they already know.

## 5. Limitations

In this study, while the robot's joints acted at their maximum pre-determined speed, these physical constraints coupled with the time-consuming nature of the necessary computations, resulted in some delays in its reactions. Additionally, due to the restricted speed of the robot's movement, the sound source may have moved out of the robot's range of vision (even though we have captured the -15/15 degrees photos). Consequently, in this case, the subject may disappear from the robot's visibility, causing the robot to be unable to locate the source of the sound. This situation arises because the proposed model always needs to detect something in the environment, leading to the robot mistakenly detecting the wrong object.

Furthermore, this fact was observed that sound reflection could confuse the robot in detecting the origin of the sound due to the unreliability of the sound localization capability of the robot. This issue can be solved by enhancing the sound signal by implementing filters or by increasing the number of microphones with better distinction quality. This system can also be equipped with an adaptive noise-canceling algorithm to filter out background environmental sounds and focus more on sudden and loud sounds.

## 6. Conclusion

this study introduced a multi-modal system designed to emulate the fear response in humans on humanoid robots, with a particular focus on the NAO robot. The designed fear module was devised to make social robots more acceptable and change their old robotic characteristic. In sound-induced fearful situations, humans instinctively react to any scary or shocking sound that may be sudden, loud, or associated with a frightening incident or creature. Next, they attempt to locate the source of the sound and identify any potential agents involved. Finally, they visually scan their surroundings for any objects or creatures that could possibly produce the sound. To replicate these reactions in robots, a preliminary exploratory study was conducted to use the human senses to imitate human reactions when hearing a sudden and loud sound. For this purpose, three primary modules were designed for sound detection and localization, object detection, and motion generation. For generating human-like arm movements, three points on each arm were chosen, and a dataset was gathered by recording RGB videos and utilizing the MediaPipe pose tracker to extract the selected points' 3D positions in the videos. A motion generator model consisting of an LSTM and an MDN model was developed to generate motion, a sound detection model was constructed based on the transfer learning technique, and for the object detection model, YOLOv5 was preferred. In the end, these models were integrated into a unified module implemented on the NAO robot. This module instantaneously detected the occurrence of the sound, turned to the direction of the incoming sound while performing human-like hand movements, and finally found the source of sound using both sound classification and object detection models. Finally, a survey gauging public perception of the robot's performance and the effectiveness of the fear module was conducted. Results indicated that the proposed module successfully convinced the majority of participants that the robot exhibited human-like behavior in response to sudden, loud sounds in its environment. Notably, expert participants had lower expectations from the robot compared to non-experts. This proof-of-concept research highlights the efficacy of the proposed method for increasing sociability and presents a potential breakthrough for similar endeavors in the field.